# A Comparative Study on Crime in Denver City Based on Machine Learning and Data Mining

Md. Aminur Rab Ratul, Faculty of Engineering, University of Ottawa, mratu076@uottawa.ca

*Abstract*— To ensure the security of the general mass, crime prevention is one of the most higher priorities for any government. An accurate crime prediction model can help the government, law enforcement to prevent violence, detect the criminals in advance, allocate the government resources, and recognize problems causing crimes. In order to construct any future-oriented tools, examine and understand the crime patterns in the earliest possible time is essential. In this paper, I analyzed a real-world crime and accident dataset of Denver county, USA, from January 2014 to May 2019, which containing 478,578 incidents. This project aims to predict and highlights the trends of occurrence that will in return support the law enforcement agencies and government to discover the preventive measures from the prediction rates. At first, I apply several statistical analysis supported by several data visualization approaches. Then, I implement various classification algorithms such as Random Forest, Decision Tree, AdaBoost Classifier, Extra Tree Classifier, Linear Discriminant Analysis, K-Neighbors Classifiers, and 4 Ensemble Models to classify 15 different classes of crimes. The outcomes are captured using two popular test methods: train-test split, and k-fold cross-validation. Furthermore, to evaluate the performance flawlessly, I also utilize precision, recall, F1-score, Mean Squared Error (MSE), ROC curve, and paired-T-test. Except for the AdaBoost classifier, most of the algorithms exhibit satisfactory accuracy. Random Forest, Decision Tree, Ensemble Model 1, 3, and 4 even produce me more than 90% accuracy. Among all the approaches, Ensemble Model 4 presented superior results for every evaluation basis. This study could be useful to raise the awareness of peoples regarding the occurrence locations and to assist security agencies to predict future outbreaks of violence in a specific area within a particular time.

I. INTRODUCTION

Crime is a colossal social issue all over the world, which damages our growth both economically and socially [1]. It contemplates many crucial factors, such as whether or not a person moves to new places and what areas they should remove from their itinerary list [2]. The United Nations Office on Drugs and Crime (UNODC) has approximated that every year, due to the violence about a half-million of people are deliberately killed [3]. 24 the deliberate and unlawful homicides occurred in per 100,000 occurrences, and some regions like Latin America has been affected mostly by endemic savagery [4]. The violence influence both the development and public health sector of a country. For example, in many countries chronic ferocity, directly and indirectly, costs up to 10% of GDP. The world's nations recently reach an agreement to lessen all kinds of acts of violence and homicides by 2030 to tackle these sterns situations [5].

To diminish the increase of outbreaks of violence and all kinds of crimes the law enforcement agencies require all sorts of futuristic automated tools that can help them to predict geographic information, types of sins, and the reasons behind the acts to protect the society and upgrade the crime analytics [6]. Crime generally has temporal and spatial attributes related to the environment, location, time, people, economic factors, social events, politics, etc. The victims of violence may not be forecasted; however, the criminal, location, occasion, and time can be predicted before. Data mining is one of the handiest tools right now to acquire meaningful knowledge and information from raw data [7]. Data mining automatically searches and inspects a large volume of data to find, discover, and learning the hidden patterns, structures, and trends between all attributes [8]. Moreover, it can answer and addresses many unknown queries that can not be answered through a usual reporting procedure [9].

A high accurate data mapping system needs a heavy amount of longitudinal data collection. It is hugely expensive in respect to money and time. Most of the rich countries have affluent data to predict the outcome from these data, although developing countries suffer from utilizing these techniques [10]. The major reasons behind these are the lack of skills to collect the data and make a decision from that. Therefore, we need an automated architecture that can help us to map and predict valuable information from crime data.

In this paper, I utilize the Denver crime dataset which contains criminal offenses record Denver county and seven districts of Denver from the year 2014 to 2019 [11]. The dataset is based on the National Incident Based Reporting System (NIBRS), which incorporates records of all victims of person crimes and all crimes within an occurrence. Furthermore, the dataset includes 19 different attributes and 478,578 instances. In this dataset, there are 15 separate offense categories, and 199 different offense categories exist.

Here, I utilize several different data mining techniques to detect the hidden pattern and relations between various features. Additionally, to inspect the result of data mining, I use the data visualization process to show those outcomes on graphs and heatmap. Nowadays, the traditional methods become outdated because of the vast amount of stored data and the complexities present on those data [12]. To get the ideal pattern from those data, we need a large amount of workforce and massive resources. The advancement of Machine Learning architectures proofs very proficient in this case for decision making, prediction, classification, and pattern recognition from huge varieties of data. Therefore, in this dataset, I employ different Machine Learning algorithms to classify 15 different categories of crimes in Denver. Additionally, to achieve accurate accuracy, I tried different dataset rebalanced procedures and two feature selection methods.

The remaining part of the paper structured as follows: in section 2, discuss several related works. The dataset and data visualization demonstrated in section 3. Section 4 illustrated the methodology, and then section 5 obtained the performance analysis and outcome of the experiments. Finally, the conclusion and discussion incorporated in section 6.

## II. RELATED WORK

In this section, I briefly discuss several methods of crime prediction. Previously most of the methods try to identify crime hotspots based on the location of huge crime density. These methods did not consider crime types or time and place of the crimes. However, recently, there are many Machine Learning and Deep Learning based approach proposed in this sector.

Cheney et al. [13] utilize hotspot mapping to predict the spatial pattern of different crimes. They tried several different mapping methods such as spatial ellipses, grid thematic mapping, point mapping, kernel density estimation, and thematic mapping of geographic areas to identify the hotspot of crimes. In [14], Bogomolov et al. use the dataset, which contains the crime information from the city of Philadelphia from the year 1991-1999. This work emphasis on multi-scale complex relationships between time and space. Decision trees, Naïve Bayes, and association rules used in [15] to predict the most significant features influence the crime. Here, data was collected from the Egyptian ministry of interior from 1996 to 2012 and the dataset containing criminal's personal details like profession, age, crime types, crime areas, social class, and education level. The accuracy of these algorithms reaches nearly 92%. Solaiman et al. [16] proposed methods based on K-means and dynamic clustering algorithms to recognize the crime hotspots where the accuracy achieved 98.7% by random subspace classifier. 1000 cases of Real-time data collected from the police department of Kuwait and the crime types are divided from several different classes such as assault, adultery, drug, forging, suicide, etc. In [17], Zubi et al. collected 350 crime records with seven different features from the police department of Libya to help the Libyan government to predict the crime patterns. In this study, the used k-means and the Apriori algorithm. To train any Machine Learning models, at first we need a decent amount of data. In the study [17, 18], I think the amount of data is not well enough to train and find good accuracy from any Machine Learning model. In [18], Almanie et al. worked with two different real-world datasets namely Denver city data (19 attributes with 333068 cases) and Los Angeles data (14 attributes with 243750 instances). They provide several statistical analyses by several graphs and utilized Naïve Bayesian classifier, Decision Tree classifier, and Apriori algorithm to classify different crime types. Here, they only use four attributes such as occurrence time, district id, geographical location, and neighborhood id to classify different classes of crimes. I believe only four attributes are not well enough to build a perfect model from any real-world crime dataset because several other relations can be established from other attributes. Ahmed and his team used the Naïve Bayes algorithm and rapid miner data mining tool to predict different types of crimes [19]. Here, they capture data from the Indian government website of the year 2012, 2013, and 2014. In [20], Jangra et al. used the Naïve Bayes classifier and K-Nearest Neighbors to predict the crime from their dataset. They showed that the Naïve Bayes classifier achieved 96.48% accuracy compared to 77.18% accuracy of KNN. Prisoners data from 2011 to 2013 on Labuan Deli prison utilized in [21] with the Decision tree algorithm. In this study, Rapidminer with a Decision tree algorithm used to analyze the data. In [22], the research team investigate Vancouver crime data for the last 15 years and used the machine learning approach to predict the crime. They mainly implement K-nearest neighbor and boosted decision tree algorithms on this dataset and achieved accuracy between 39% to 44%. In my perspective, the accuracy is too low to build any practical life applications.

Kang et al. [23] proposed a deep neural network for crime prediction by merging multi-modal data from different domains. The dataset contains 274,064 cases of 2014 and the proposed model displayed accuracy nearly 84.25%. Another Deep neural network based model presented in [24], where Stec et al. predict next day crime counts after employ their model on Chicago and Portland datasets. They build several models but their best model achieves 75.6% and 65.3% accuracy respectively on Chicago and Portland datasets. Wang et al. [25] provide a deep learning approach to predict real-time crime forecasting. Here, the authors firstly offer a perfect representation of crime data. After that, they employ a spatial-temporal residual network on this representational dataset to forecast the crime distribution of Los Angeles at the scale of hours. Duan and his team presented a novel Spatiotemporal Crime Network (STCN) with a deep Convolutional Neural Networks (CNNs) to accomplished crime-referenced feature extraction [26]. This research team utilizes data in New York City from 2010 to 2015 and predicts the next day risk of crime in each region's urban area. This Spatiotemporal Crime Network (STCN) achieve an 88% F-1 score and 92% AUC. Huang et al. [27] proposed an attentive hierarchical recurrent networks based approach called DeepCrime for crime prediction. This architecture carefully detects the inter-dependencies between crimes and other data in urban space to exhibit the crime patterns.

## III. DATASET AND DATA VISUALIZATION

### A. Dataset

In this paper, I used a real-world Denver crime dataset, which contains 478,578 instances and 19 attributes in this dataset. It includes the crime incidents and offenses of crimes of Denver county for the last five years and also to the running year (January 2014 – May 2019). The key attributes of this dataset are offense type and category of offenses. Furthermore, many essential features also existed in this dataset, such as the first occurrence date and time of the crime, the reported date and time of the crime, the exact geographical location of the crime, the neighborhood information, the exact place where the violence happened, etc. In table 1, I provide the details description of all attributes of the Denver crime dataset.

TABLE I. ATTRIBUTES DESCRIPTION OF DENVER CRIME DATASET

| Attribute | Attribute Description | Number of Values |
|---|---|---|
| INCIDENT_ID | unique identifier for an occurrence of offenses (unique id for the incident) | Unlimited values |
| OFFENSE_ID | unique identifier for each offense which is the combining values of INCIDENT_ID, OFFENSE_CODE, and | Unlimited values |

| | OFFENSE_CODE_EXTENSION | |
|---|---|---|
| OFFENSE_CODE | Unique identifier for a particular type of offense | Unlimited values |
| OFFENSE_CODE_EXTENSION | Extension ID of the OFFENSE_CODE | 6 different values |
| OFFENSE_TYPE_ID | Provide exact name of the offense | 15 different values |
| OFFENSE_CATEGORY_ID | Present more general categorization of crimes | 199 different values |
| FIRST_OCCURRENCE_DATE | First possible time of the occurrence of the offense | Unlimited date and time values |
| LAST_OCCURRENCE_DATE | Last possible time of the occurrence of the offense | Unlimited date and time values |
| REPORTED_DATE | Time and date when the violence reported to the police | Unlimited date and time values |
| INCIDENT_ADDRESS | Provide the street address of the offense | Unlimited address values |
| GEO_X | State plane X-axis value of the occurrence location | Unlimited values of Geocode X-axis |
| GEO_Y | State plane Y-axis value of the occurrence location | Unlimited values of Geocode Y-axis |
| GEO_LON | longitudes of crime occurrence | Unlimited geographical longitudes of the offenses |
| GEO_LAT | Latitudes of crime occurrence | Unlimited geographical latitudes of the offenses |
| DISTRICT_ID | District in-charge of the particular offense | 7 different values |
| PRECINCT_ID | Particular Precinct in charge of handling the offense | 36 different values |
| NEIGHBORHOOD_ID | The neighborhood where the offense occurred | Unlimited address values of location |
| IS_CRIME | Whether a particular offense was a criminal offense or not (0 for not, 1 for yes) | 2 different values |
| IS_TRAFFIC | Whether a particular offense was a traffic-related incident or not (0 for not, 1 for yes) | 2 different values |

*B. Data Analysis and Visualization*

To acquire proper knowledge on the dataset, I apply several statistical analyses on the attributes of this dataset. I try to map several attributes of the dataset and try to establish a relationship between them to extract some essential trends and patterns of this data. I produce several different graphs and some heatmap to visualize and understand the data neatly. Each graph provides percentages of violence occurrences regarding a particular situation. I create several varieties of graphs to draw the real picture from this dataset.

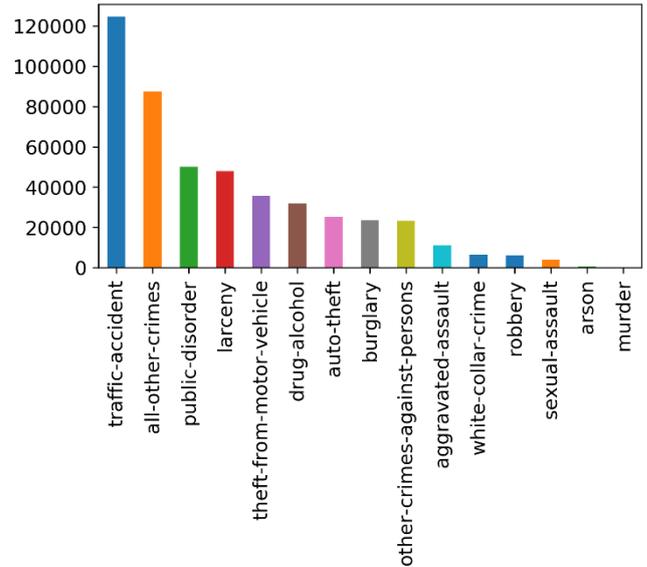

Fig. 1. 15 types of offenses of Denver Crime dataset with incidents amount

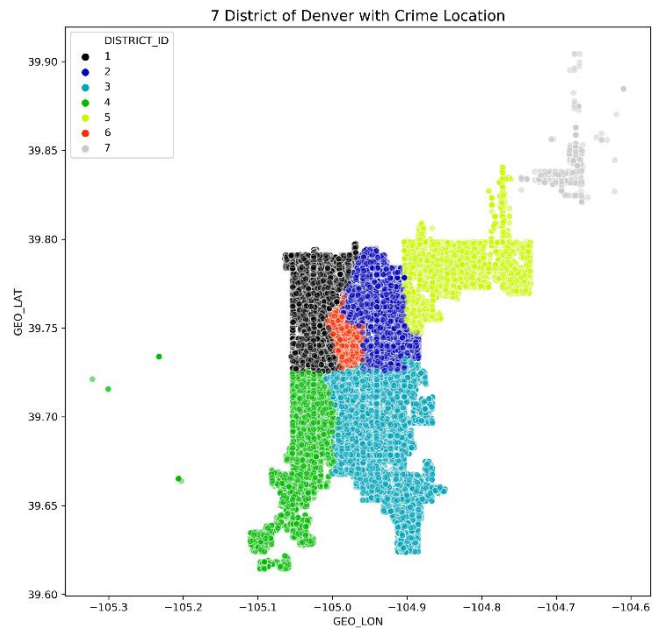

Fig. 2. Seven districts of Denver with geographical location where crime took part

Figure 1 displayed the 15 types of crimes of Denver with the exact number of cases. Traffic accident related crime is the most frequent offenses in Denver from 2014 to 2019, where more than 120,000 crimes took place, whereas the least significant crimes were arson and murder. In figure 2, I divide the 7 districts of Denver with crime location based on geographical latitudes and longitudes. In figure 3, through the bar graph, I try to exhibit the real picture of different kinds of crime in Denver. In this time frame (2014-2019), districts 1, 3, and 6 were more vulnerable to offenses. Furthermore, traffic accident related crime in regions 3 and 1 was excessively high. Arson, murder, robbery, sexual assault, and white crime were the least significant in this period.

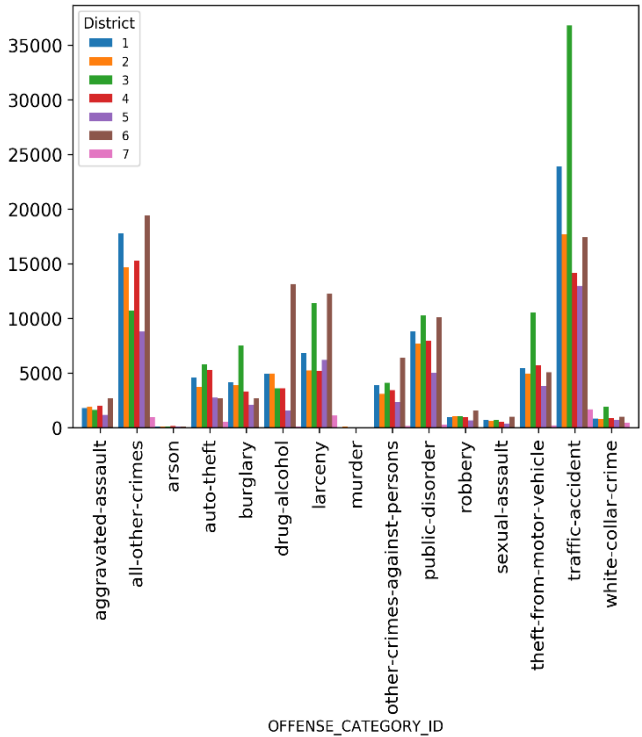

Fig. 3. 15 types of offenses of Denver crime dataset in seven districts

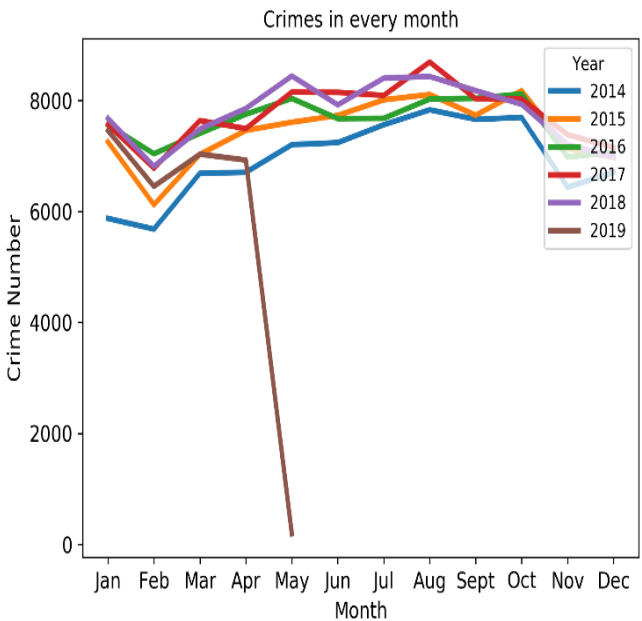

Fig. 4. Crimes rate per month from 2014 to 2019

In Figures 4 and 5, I provide the crime rate respectively per month and hour in this period (2014-2019). In this dataset, I find the data from January 2014 to May 2019, that is the reason the graph shows the data until May 2019. From this figure, I can see that every year, May to August are the most vulnerable to crime, and the violence was high in 2017 and 2018. Figure 5 reveals that 12 pm- 6 pm (12-18) was the pick time when most of the offenses occurred. Specifically, 4 pm (16) is the most unsafe time for the general mass. The crime rate get rises after 6 am (6), and it becomes low after 7 pm (19); however, many crimes are showing an upward trend after 8 pm (20). 3 am to 5 am are the most secure time in a day. I presented a bar graph in figure 6 that shows on a particular day which hour getting the most traffic offenses. The frequency of traffic crime is most from 1 pm to 6 pm (13-18). Though 3 am to 5 am is the most secure time in terms of traffic crimes.

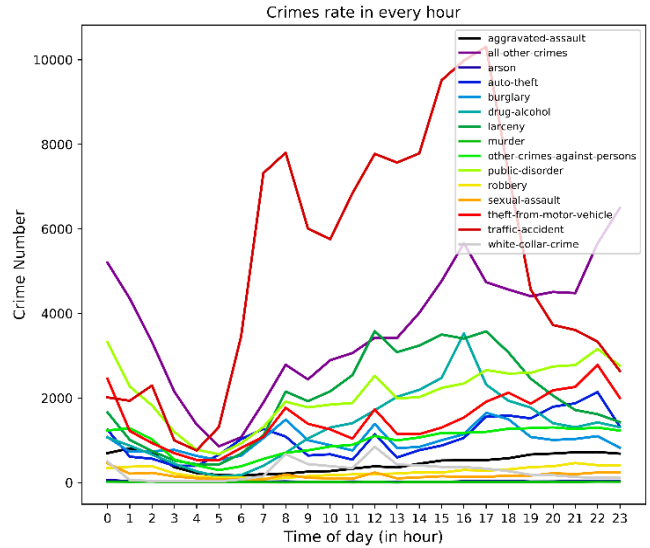

Fig. 5. 15 kinds of crime rate per hour from 2014 to 2019

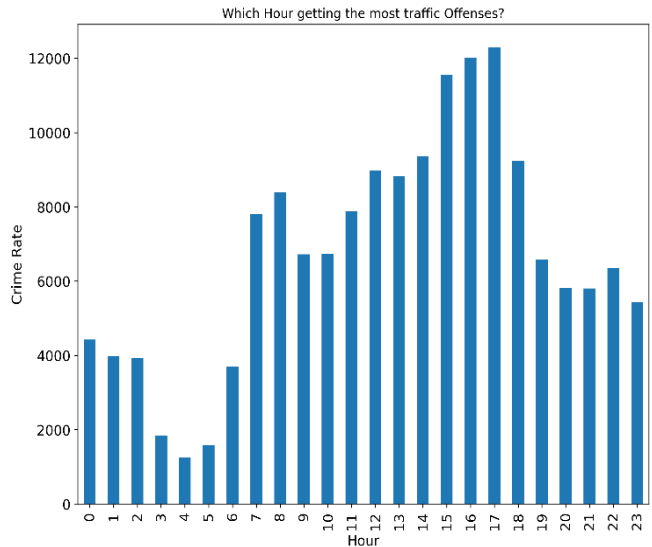

Fig. 6. Which hour more vulnerable to traffic crime (year: 2014-2019)

In figure 7, I disclose the average crime complain rate per day in these six years. Moreover, Christmas day (25th December) is the safest time of the year, although June 1st was the most unsafe time. The first few days of each month and some middle part of the month are more vulnerable to the misdeed. According to this pivot table, Summer (May-August) and the early Fall (September and October) are the riskiest time of the year. Specifically, August is the most unsafe month of the year. On the other hand, Winter (December to March) can be safe in Denver city and county.

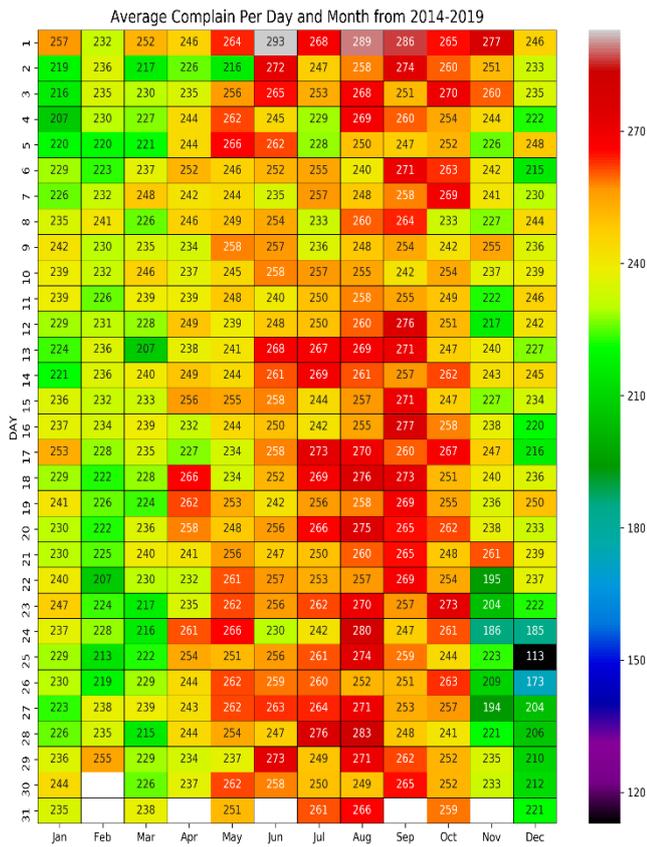

Fig. 7. Average complaint rate per day from 2014 to 2019

From Figures 1 and 3, I get an intuition that traffic accident linked crime is most frequent in Denver in this period, and district 3 is the most unprotected place for this activity. Thus, in figure 8, I attempt to exhibit the traffic accident associated crime through a snapshot of a heatmap. In figure 9, I provide a snapshot of a heatmap which unveil specific locations of the crime when there is also traffic jam coexist at the same time in these 6 years time span.

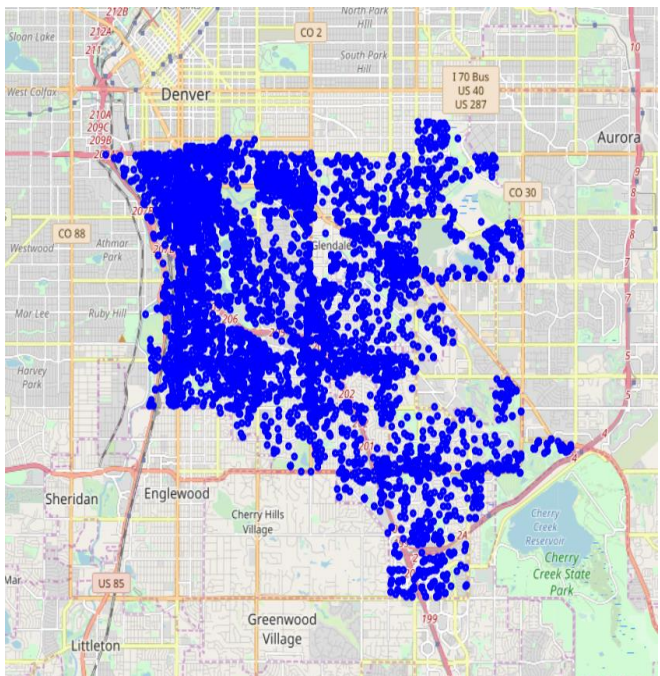

Fig. 8. Traffic accident associated crimes in district 3 from 2014-2019

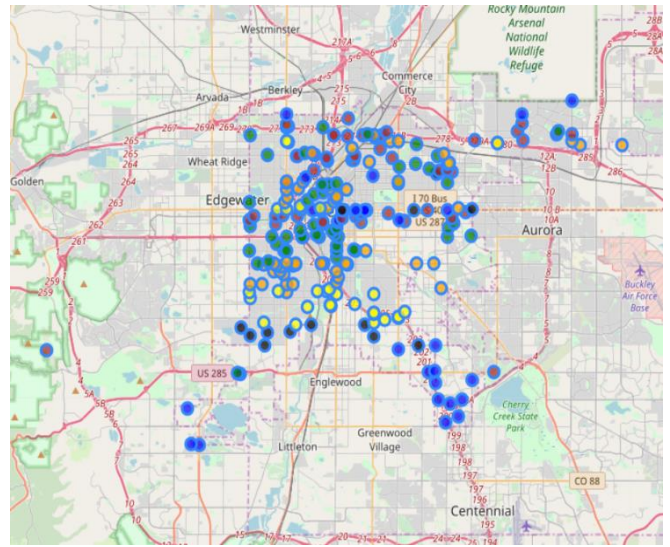

Fig. 9. Specific locations of the crime when there is also traffic jam coexist

## IV. METHODOLOGY

I strongly believe that discover the relationship between the attributes is so valuable to detecting critical hotspots for a futuristic automated model [18, 28]. I anticipate this type of crime prediction method always helps police, law enforcement authorities, security agencies, and the public to forecast the real crime incidents before it is happened not only in Denver but also in all over the world.

In this paper, I attempt to classify 15 different categories of offenses in Denver with respect to the exact time and specific location. From the previous part from the data visualization, I get the intuition that which features are important to find the exciting patterns among all the variables in this crime dataset. Then I divide my proposed methodology mainly in four parts: data acquisition, data pre-processing, building a perfect automated classification system after comparing several algorithms, and at last reach an ideal decision and conclusion from the previous stage.

### A. Data Acquisition

I already described in the previous section that I collected real-world crime data from Denver county and city from January 2014 to May 2019 [11]. This dataset the 478,578 instances and 19 attributes.

### B. Data Pre-Processing

*a) Data Cleaning:* There are many missing values, particularly in two attributes in the Denver crime dataset: incident_address and last_occurance_date [18]. At first, I put a huge negative value such as -9999 when I find an empty place in my data frame. However, these two attributes are not important or key attributes for my experiment. Therefore, I drop these two attributes from my data frame, and I did not need to clean these. All the key attributes in this dataset are not empty and these were complete. Additionally, I did not require to clean them and I observe that these attributes did not have any inconsistent and noisy values.

*b) Data Reduction:* To select only vital attributes for our training phase, I apply the data reduction technique in this dataset. I employ the dimensionality reduction procedure by utilizing attribute subset selection. Among 19 attributes, I selected 13 attributes and remove some irrelevant ones like first occurrence date, last occurrence date, incident id, incident address, offense id, and offense code extension.

*c) Data Integration:* For prediction and reach any decision from the crime dataset, I believe reported date and time can be a crucial part. So, I convert crime reported date attribute to 4 new attributes such as year, month, day, and hour. I apply the Military time system to achieve these 4 new attributes and to get more frequent patterns. I only take the hours values and remove minutes and seconds from my consideration. Therefore, the final dimension of my dataset is (478577 × 16).

*d) Data Conversion:* In this part of my experiment, I convert my object type data into categorical data. Before feeding the data to the architectures, it is an important process to change object data to categorical.

*e) Data Shuffling:* In this stage, I shuffle my dataset. Data shuffling crucial for machine learning algorithms because the main focus of this functionality to reduces the variance, model remain general and overfit less. Thus, it helps to improve the accuracy of my models.

*f) Data Normalization:* Normalization is an essential data preparation technique. I used min-max normalization methods to normalize my data and give my data value between 0 to 1 so that my data will not face any distorting distinct ranges of values. It is necessary for my dataset because, for most of the features, different ranges exist.

$$nv = (\frac{v - \min(A)}{\max(A) - \min(A)}(1 - 0) + 0) \quad (1)$$

In this equation (1) $nv$ = new value; $v$ = particular value from that specific attribute $A$; $\min(A)$ = minimum value of this attribute $A$, $\max(A)$ = max value of this attribute $A$; normalized range = (0 to 1)

*g) Data Sampling:* I used several data sampling methods during the training process of my algorithms to inspect that which methods give me better accuracy. To oversample the data, I used two different techniques, such as random oversampling and SMOTE oversampling. Besides, to undersample the data, two of my methods are random undersample and TOM-LINK undersampling method. Finally, for balanced data sampling, I apply SMOTETomek balanced sampling method.

*h) Feature Selection:* Here, I utilized two different feature selection methods after getting the result from the rebalance dataset or data sampling part. First of all, I use the SelectKBest feature selection method with the ANOVA technique. Secondly, I used the VarianceThreshold method with a threshold value (0.8 * (1 - 0.8)).

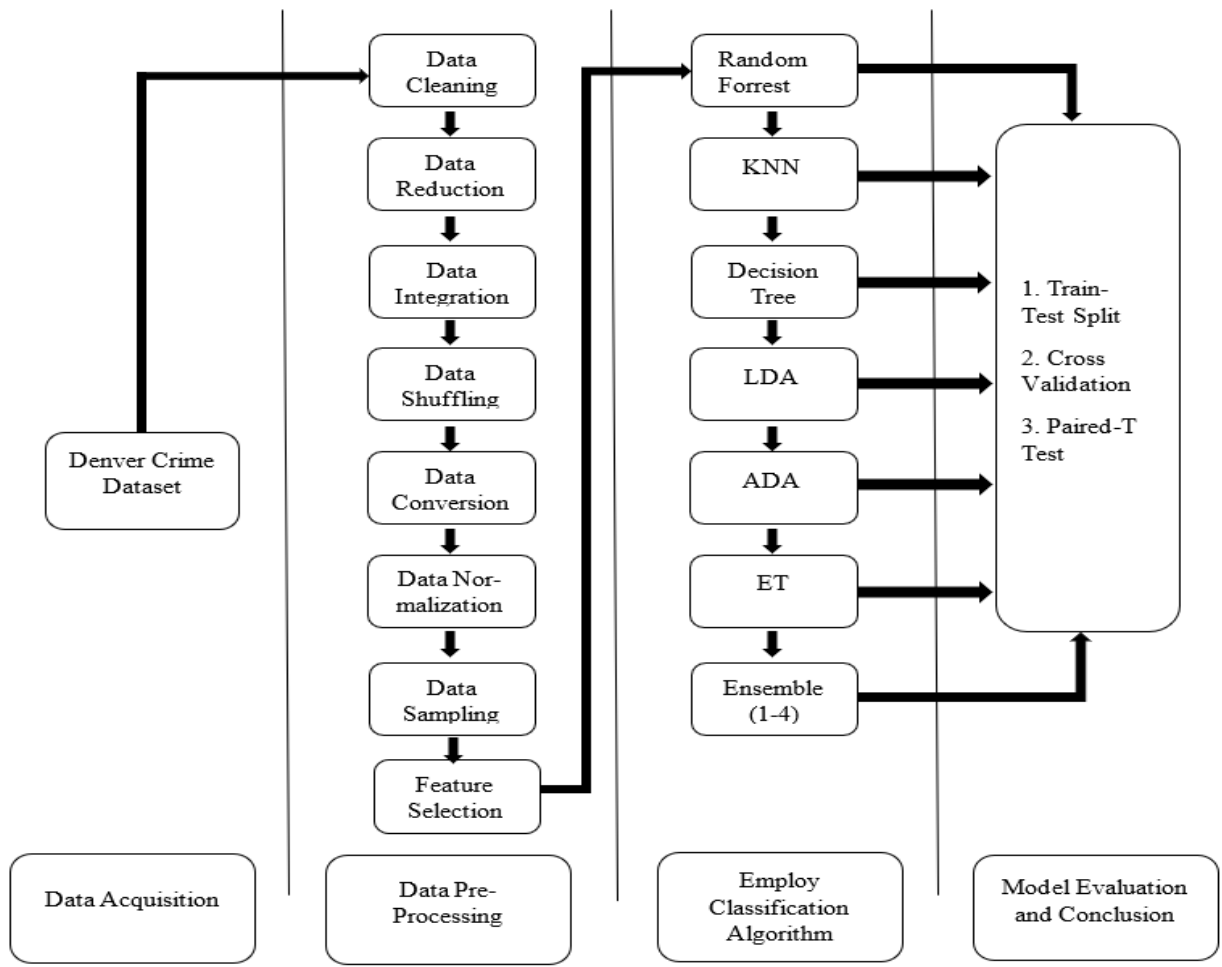

Fig. 10. Step by step implementation process of my proposed methodology

## C. Model Building

After the data pre-processing part, I apply several available machine learning algorithms on my newly created data frame to classify 15 different types of crimes. After implementing and check the performance of all models, I proposed five ensemble models for this dataset, which help many low performed methods to improve the result. The primary purpose of the classification task is to build a future-oriented model that can detect distinct types of crimes in a particular location within a specific time. Here, I inspect every model, and then I will choose which model gives me the best outcome. In this section, I describe all the algorithms I utilized.

*a) Random Forest [29]:* It is a meta estimator which fits several numbers of decision trees on a diverse sub-samples of the dataset and utilize averaging to enhance the accuracy of the proposed model. In addition, try to limit the overfitting. Since the generated trees were complex, I trained the random forest with some different the max tree depth ranges and several different trees range from 10 to 200. However, after considering many important issues like overfitting, accuracy, Mean Squared Error (MSE), and underfitting, I decide to use Max depth=7, Number of trees=100. Additionally, I apply "entropy" to split and information gain.

*b) Decision Tree Classifier [30]:* It is a non-parametric classification model that predicts the outcome of a target variable by learning many simple decision rules implied from the data features. Though the decision tree is an uncomplicated portrayal of knowledge, it can be handy to solve any easy practical life solutions to complicated ones [31]. To implement the decision tree classifier, I use the maximum tree leaf size is 7, to avoid the overfitted model. Moreover, for the information gain, I used "entropy" to maintain the quality of a split. I check accuracy and MSE to decide which decision tree is perfect for the dataset as mentioned above.

*c) K-Neighbors Classifiers (KNN):* It is an instance-based and non-parametric learning algorithm. It makes the future predictions from all the training data by calculating the similarities between an input sample and each training example. I employ 5 as the K-Neighbors and weights are 'uniform' so that all points in each neighborhood get equal weight. Lastly, I apply euclidean distance to calculate the distance for the attributes.

*d) Linear Discriminant Analysis (LDA):* It is a classifier with Linear decision boundaries. This classifier utilizing Bayes rules and initiated by adjusting class conditional densities to the existing data. As my dataset has a decent amount of features so, I apply Singular value decomposition (SVD) as the solver.

*e) AdaBoost Classifier (ADA) [32]:* ADA is a meta-estimator classifier that starting by fitting a classifier on the primary dataset and then fits extra copies of the classifier on the very same dataset. I operate this algorithm with n_estimators = 50, at which boosting will be terminated.

*f) ExtraTrees Classifier (ET):* ET is a meta-estimator classifier that implements several numbers of decision trees (extra-trees) on diverse sub-samples of the dataset. Then calculate the average value to increases the predictive accuracy and minimizes overfitting. For this classifier, I use 100 trees and maximum depth = 7 for trees. Furthermore, for this classifier, I apply 'gini' or Gini impurity to split.

*g) Ensemble Model 1:* For ensemble model 1, I apply three different algorithms as estimators: 1) Random Forest, 2) Decision Tree Classifier, and 3) Linear Discriminant Analysis (LDA). In addition, I apply the majority voting methodology ("hard" voting) to implement this ensemble model.

*h) Ensemble Model 2:* I implement my second ensemble model based on Bagging Classifier. I used three algorithms from ensemble model 1 (Random Forest, Decision Tree Classifier, and Linear Discriminant Analysis (LDA)) to implement this one. Basically, Bagging Classifier is an ensemble meta-estimator that adjusts each base classifier on the random subsets of the main dataset. Then to achieve the final prediction I have to aggregate their value by voting or by averaging. I use max features = 0.5 and max samples = 0.5 to train this model. Max features and max sample parameters are drawn from the data frame to train each base estimators..

*i) Ensemble Model 3:* For the 3rd Ensemble method, I apply the Random Forest algorithm with Extra Tree classifier (ET) and K-Neighbors Classifier (KNN). To boost the performance of KNN and ET, I applied them with the Random Forest classifier (because in the result section, I will show that the random forest gives me high accuracy). I implemented this ensemble model with the "soft" voting technique, which means it predicts the class labels based on the argmax of the sums of predicted probabilities. I put particular weight on different algorithms to build this model. Weight 1, 2, 2, respectively on Random Forest, KNN, and ET. The reason behind to add extra weight on two low performers to improve their performance. This less weight in a high performer (Random Forest) can relieve the model from overfitting. When I provide a different weight on different algorithms, each classifier predicted class probabilities are collected. After that multiply by the weight of classifier which I offered, and then calculate the average. The final class labels then attain from the class label with the superior average probability.

*j) Ensemble Model 4:* For my fourth and last ensemble model, I selected three tree-based models with bagging and voting. Firstly, I pick three based classifiers (Random Forest, Decision Tree, Extra Tree Classifier) for this architecture. Secondly, I employ the Bagging Classifier method on each of these three algorithms and acquired three new results. Finally, I apply the Voting Classifier method on previously found three Bagging Classifier outputs and achieve the final accuracy. Like earlier, for Bagging Classifier, I used max samples=0.5, max features=0.5. Additionally, for voting, I apply the majority voting criteria ("hard" voting). Figure 11 displayed the details steps of the 4th ensemble model.

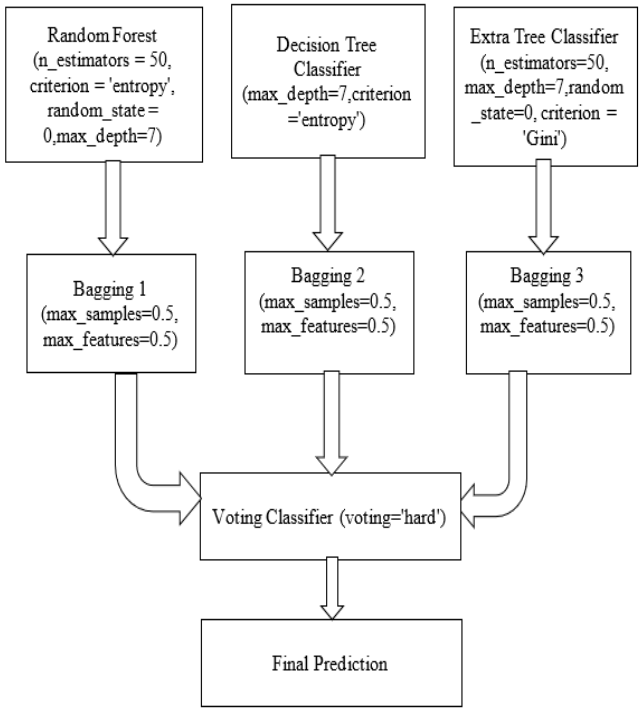

Fig. 11. Each Step of Ensemble Model 4

## V. EXPERIMENTAL RESULT

### A. Practical Implementation

For practical implementation, I used several python libraries. For instance, at first for data pre-processing, I used pandas, and to visualize my data correctly, I used Mattplotlib, Seaborn. Furthermore, I employ the folium library to produce the Heatmap and at last utilized Sklearn for training, testing, and evaluate all the algorithms. This work performed on Intel Core i7-8750H with 4.1 GHz and an NVIDIA GeForce GTX 1050Ti GPU with 4GB GDDR5 dedicated VRAM.

### B. Model Evaluation

The primary purpose of this work to classify 15 different types of crimes and established a future-oriented automated model. To evaluate my models, I apply four evaluation criteria to judge their performance and accuracy. Firstly, I executed the k-fold cross-validation for ten folds. Then to check my models further, I performed a train-test split method with 76% data for training and 34% data for testing. After that, I provide the Mean-Squared-Error (MSE) to check the error value of the algorithms produce. Then, produce the Paired-T Test result to verify that the algorithms which give me better accuracy, they are significantly different or not. Lastly, I demonstrate the ROC curve to examine the result of each class.

*a) K-Fold Cross Validation:* In this part, I will display the 10 folds cross-validation outcome for all of my models. In table 2-11, I provide the cross-validation result before and after data sampling for all of my models. To 10 folds divide, I use Stratified-KFold cross-validation to retain the same ratio between class in training and testing. Here I produce the result for cross-validation results of normal cross-validation, random oversampling, random undersampling, SMOTE oversampling, TomeLinks undersampling, and SMOTETomek balanced sampling for every model.

TABLE II. CROSS-VALIDATION RESULT FOR RANDOM FOREST

| Folds | Cross-validation (Before Sampling) | Random Oversample | Random Undersample | SMOTE Oversample | Tome Links Under sample | SMOTE Tomek Balanced Sample |
|---|---|---|---|---|---|---|
| 1 | 0.66 | 0.92 | 0.90 | 0.91 | 0.82 | 0.91 |
| 2 | 0.66 | 0.92 | 0.93 | 0.90 | 0.81 | 0.89 |
| 3 | 0.67 | 0.91 | 0.88 | 0.87 | 0.85 | 0.89 |
| 4 | 0.66 | 0.92 | 0.93 | 0.89 | 0.83 | 0.92 |
| 5 | 0.70 | 0.91 | 0.89 | 0.89 | 0.82 | 0.93 |
| 6 | 0.69 | 0.91 | 0.93 | 0.85 | 0.82 | 0.92 |
| 7 | 0.72 | 0.92 | 0.89 | 0.87 | 0.83 | 0.89 |
| 8 | 0.74 | 0.89 | 0.88 | 0.85 | 0.82 | 0.89 |
| 9 | 0.72 | 0.92 | 0.88 | 0.88 | 0.83 | 0.92 |
| 10 | 0.71 | 0.93 | 0.88 | 0.88 | 0.84 | 0.91 |
| Accuracy | 0.69 | 0.92 | 0.90 | 0.88 | 0.83 | 0.91 |

TABLE III. CROSS-VALIDATION RESULT FOR DECISION TREE

| Folds | Cross-validation (Before Sampling) | Random Oversample | Random Undersample | SMOTE Oversample | Tome Links Under sample | SMOTE Tomek Balanced Sample |
|---|---|---|---|---|---|---|
| 1 | 0.99 | 0.98 | 0.98 | 0.97 | 0.98 | 0.99 |
| 2 | 0.99 | 0.97 | 0.97 | 0.95 | 0.98 | 0.99 |
| 3 | 0.99 | 0.98 | 0.98 | 0.96 | 0.97 | 0.99 |
| 4 | 0.99 | 0.97 | 0.97 | 0.96 | 0.97 | 0.99 |
| 5 | 0.99 | 0.97 | 0.97 | 0.97 | 0.96 | 0.99 |
| 6 | 0.99 | 0.98 | 0.98 | 0.95 | 0.98 | 0.99 |
| 7 | 0.99 | 0.97 | 0.98 | 0.97 | 0.98 | 0.99 |
| 8 | 0.99 | 0.97 | 0.99 | 0.96 | 0.97 | 0.99 |
| 9 | 0.99 | 0.96 | 0.98 | 0.96 | 0.98 | 0.99 |
| 10 | 0.99 | 0.97 | 0.98 | 0.96 | 0.98 | 0.99 |
| Accuracy | 0.99 | 0.97 | 0.98 | 0.96 | 0.98 | 0.99 |

TABLE IV. CROSS-VALIDATION RESULT FOR LINEAR DISCRIMINANT ANALYSIS (LDA)

| Folds | Cross-validation (Before Sampling) | Random Oversample | Random Undersample | SMOTE Oversample | Tome Links Under sample | SMOTE Tomek Balanced Sample |
|---|---|---|---|---|---|---|
| 1 | 0.73 | 0.14 | 0.73 | 0.76 | 0.73 | 0.14 |
| 2 | 0.73 | 0.14 | 0.74 | 0.76 | 0.73 | 0.14 |
| 3 | 0.73 | 0.14 | 0.72 | 0.76 | 0.73 | 0.14 |
| 4 | 0.73 | 0.14 | 0.73 | 0.76 | 0.73 | 0.14 |
| 5 | 0.73 | 0.14 | 0.72 | 0.76 | 0.73 | 0.14 |

| | | | | | |
|---|---|---|---|---|---|
| 6 | 0.73 | 0.14 | 0.71 | 0.76 | 0.73 | 0.14 |
| 7 | 0.73 | 0.14 | 0.71 | 0.76 | 0.73 | 0.14 |
| 8 | 0.73 | 0.72 | 0.73 | 0.76 | 0.73 | 0.73 |
| 9 | 0.73 | 0.14 | 0.72 | 0.76 | 0.73 | 0.14 |
| 10 | 0.73 | 0.14 | 0.73 | 0.76 | 0.73 | 0.14 |
| Accuracy | 0.73 | 0.20 | 0.72 | 0.76 | 0.73 | 0.20 |

TABLE V. CROSS-VALIDATION RESULT FOR LINEAR K-NEIGHBORS CLASSIFIERS (KNN)

| Folds | Cross-validation (Before Sampling) | Random Oversample | Random Undersample | SMOTE Oversample | Tome Links Undersample | SMOTE Tomek Balanced Sample |
|---|---|---|---|---|---|---|
| 1 | 0.77 | 0.92 | 0.53 | 0.82 | 0.77 | 0.81 |
| 2 | 0.77 | 0.92 | 0.51 | 0.82 | 0.77 | 0.82 |
| 3 | 0.78 | 0.92 | 0.53 | 0.82 | 0.78 | 0.83 |
| 4 | 0.77 | 0.93 | 0.52 | 0.82 | 0.77 | 0.84 |
| 5 | 0.78 | 0.94 | 0.50 | 0.82 | 0.78 | 0.84 |
| 6 | 0.78 | 0.94 | 0.47 | 0.82 | 0.78 | 0.84 |
| 7 | 0.77 | 0.94 | 0.47 | 0.82 | 0.77 | 0.84 |
| 8 | 0.77 | 0.94 | 0.47 | 0.82 | 0.77 | 0.84 |
| 9 | 0.77 | 0.95 | 0.50 | 0.82 | 0.77 | 0.84 |
| 10 | 0.78 | 0.94 | 0.51 | 0.82 | 0.78 | 0.84 |
| Accuracy | 0.77 | 0.94 | 0.50 | 0.82 | 0.77 | 0.84 |

TABLE VI. CROSS-VALIDATION RESULT FOR EXTRA TREE CLASSIFIERS (ET)

| Folds | Cross-validation (Before Sampling) | Random Oversample | Random Undersample | SMOTE Oversample | Tome Links Undersample | SMOTE Tomek Balanced Sample |
|---|---|---|---|---|---|---|
| 1 | 0.80 | 0.90 | 0.88 | 0.82 | 0.80 | 0.90 |
| 2 | 0.80 | 0.89 | 0.88 | 0.82 | 0.80 | 0.89 |
| 3 | 0.81 | 0.87 | 0.88 | 0.83 | 0.81 | 0.89 |
| 4 | 0.81 | 0.88 | 0.90 | 0.81 | 0.81 | 0.89 |
| 5 | 0.82 | 0.89 | 0.91 | 0.82 | 0.82 | 0.90 |
| 6 | 0.81 | 0.88 | 0.89 | 0.82 | 0.81 | 0.90 |
| 7 | 0.80 | 0.89 | 0.89 | 0.83 | 0.80 | 0.90 |
| 8 | 0.80 | 0.87 | 0.92 | 0.82 | 0.80 | 0.88 |
| 9 | 0.82 | 0.89 | 0.92 | 0.84 | 0.82 | 0.88 |
| 10 | 0.81 | 0.88 | 0.88 | 0.83 | 0.81 | 0.88 |
| Accuracy | 0.81 | 0.88 | 0.90 | 0.82 | 0.81 | 0.89 |

TABLE VII. CROSS-VALIDATION RESULT FOR ADABOOST CLASSIFIERS (ADA)

| Folds | Cross-validation (Before Sampling) | Random Oversample | Random Undersample | SMOTE Oversample | Tome Links Undersample | SMOTE Tomek Balanced Sample |
|---|---|---|---|---|---|---|
| 1 | 0.45 | 0.2 | 0.2 | 0.41 | 0.45 | 0.20 |
| 2 | 0.45 | 0.2 | 0.2 | 0.41 | 0.45 | 0.20 |
| 3 | 0.45 | 0.2 | 0.2 | 0.41 | 0.45 | 0.20 |
| 4 | 0.45 | 0.2 | 0.2 | 0.41 | 0.45 | 0.20 |
| 5 | 0.45 | 0.2 | 0.2 | 0.41 | 0.45 | 0.20 |
| 6 | 0.45 | 0.2 | 0.2 | 0.41 | 0.45 | 0.20 |
| 7 | 0.45 | 0.2 | 0.2 | 0.41 | 0.45 | 0.20 |
| 8 | 0.45 | 0.2 | 0.2 | 0.41 | 0.45 | 0.20 |
| 9 | 0.45 | 0.2 | 0.2 | 0.41 | 0.45 | 0.20 |
| 10 | 0.45 | 0.2 | 0.2 | 0.41 | 0.45 | 0.20 |
| Accuracy | 0.45 | 0.2 | 0.2 | 0.41 | 0.45 | 0.20 |

TABLE VIII. CROSS-VALIDATION RESULT FOR ENSEMBLE MODEL 1

| Fold | Normal Cross-validation | Random Oversample | Random Undersample | SMOTE Oversample | Tome Links Undersample | SMOTE Tomek Balanced Sample |
|---|---|---|---|---|---|---|
| 1 | 0.92 | 0.97 | 0.96 | 0.92 | 0.92 | 0.97 |
| 2 | 0.90 | 0.97 | 0.96 | 0.93 | 0.90 | 0.97 |
| 3 | 0.92 | 0.97 | 0.95 | 0.94 | 0.92 | 0.98 |
| 4 | 0.90 | 0.97 | 0.95 | 0.92 | 0.90 | 0.97 |
| 5 | 0.92 | 0.97 | 0.94 | 0.93 | 0.92 | 0.97 |
| 6 | 0.91 | 0.97 | 0.95 | 0.93 | 0.91 | 0.97 |
| 7 | 0.90 | 0.97 | 0.96 | 0.92 | 0.90 | 0.97 |
| 8 | 0.93 | 0.97 | 0.95 | 0.93 | 0.93 | 0.97 |
| 9 | 0.92 | 0.97 | 0.96 | 0.93 | 0.92 | 0.97 |
| 10 | 0.92 | 0.96 | 0.96 | 0.92 | 0.92 | 0.97 |
| Accuracy | 0.92 | 0.97 | 0.95 | 0.93 | 0.91 | 0.97 |

TABLE IX. CROSS-VALIDATION RESULT FOR ENSEMBLE MODEL 2

| Fold | Normal Cross-validation | Random Oversample | Random Undersample | SMOTE Oversample | Tome Links Undersample | SMOTE Tomek Balanced Sample |
|---|---|---|---|---|---|---|
| 1 | 0.63 | 0.70 | 0.71 | 0.75 | 0.70 | 0.71 |
| 2 | 0.68 | 0.74 | 0.69 | 0.76 | 0.72 | 0.71 |
| 3 | 0.71 | 0.73 | 0.70 | 0.74 | 0.62 | 0.65 |
| 4 | 0.72 | 0.73 | 0.65 | 0.75 | 0.70 | 0.73 |
| 5 | 0.71 | 0.74 | 0.72 | 0.77 | 0.70 | 0.74 |
| 6 | 0.70 | 0.71 | 0.67 | 0.73 | 0.70 | 0.73 |
| 7 | 0.61 | 0.60 | 0.71 | 0.63 | 0.70 | 0.70 |

| 8 | 0.70 | 0.72 | 0.73 | 0.72 | 0.73 | 0.74 |
| 9 | 0.73 | 0.69 | 0.70 | 0.77 | 0.66 | 0.71 |
| 10 | 0.65 | 0.73 | 0.72 | 0.73 | 0.62 | 0.73 |
| Accuracy | 0.68 | 0.71 | 0.70 | 0.73 | 0.69 | 0.71 |

TABLE X. CROSS-VALIDATION RESULT FOR ENSEMBLE MODEL 3

| Folds | Cross-validation (Before Sampling) | Random Oversample | Random Undersample | SMOTE Oversample | TomeLinks Undersample | SMOTETomek Balanced Sample |
|---|---|---|---|---|---|---|
| 1 | 0.82 | 0.90 | 0.93 | 0.89 | 0.83 | 0.90 |
| 2 | 0.83 | 0.90 | 0.88 | 0.90 | 0.83 | 0.93 |
| 3 | 0.82 | 0.90 | 0.92 | 0.88 | 0.84 | 0.89 |
| 4 | 0.82 | 0.92 | 0.89 | 0.90 | 0.83 | 0.93 |
| 5 | 0.82 | 0.91 | 0.91 | 0.87 | 0.82 | 0.92 |
| 6 | 0.82 | 0.94 | 0.91 | 0.89 | 0.83 | 0.92 |
| 7 | 0.83 | 0.91 | 0.91 | 0.90 | 0.83 | 0.92 |
| 8 | 0.83 | 0.91 | 0.93 | 0.92 | 0.83 | 0.90 |
| 9 | 0.83 | 0.92 | 0.89 | 0.89 | 0.82 | 0.89 |
| 10 | 0.83 | 0.91 | 0.88 | 0.91 | 0.82 | 0.88 |
| Accuracy | 0.82 | 0.91 | 0.91 | 0.90 | 0.83 | 0.91 |

TABLE XI. CROSS-VALIDATION RESULT FOR ENSEMBLE MODEL 4

| Fold | Normal Cross-validation | Random Oversample | Random Undersample | SMOTE Oversample | TomeLinks Undersample | SMOTETomek Balanced Sample |
|---|---|---|---|---|---|---|
| 1 | 0.95 | 0.96 | 0.97 | 0.92 | 0.96 | 0.97 |
| 2 | 0.95 | 0.97 | 0.95 | 0.90 | 0.94 | 0.98 |
| 3 | 0.96 | 0.96 | 0.97 | 0.89 | 0.92 | 0.97 |
| 4 | 0.97 | 0.98 | 0.98 | 0.90 | 0.95 | 0.97 |
| 5 | 0.93 | 0.97 | 0.96 | 0.97 | 0.90 | 0.96 |
| 6 | 0.92 | 0.98 | 0.96 | 0.91 | 0.92 | 0.93 |
| 7 | 0.96 | 0.98 | 0.97 | 0.95 | 0.92 | 0.98 |
| 8 | 0.92 | 0.98 | 0.98 | 0.95 | 0.96 | 0.98 |
| 9 | 0.90 | 0.97 | 0.97 | 0.87 | 0.93 | 0.97 |
| 10 | 0.96 | 0.96 | 0.98 | 0.94 | 0.90 | 0.98 |
| Accuracy | 0.94 | 0.97 | 0.97 | 0.92 | 0.93 | 0.97 |

*b) Train-Test Split, Feature Selection, And MSE:*

In this section, I exhibit the output for the train-test split, two feature selection methods, and mean squared error (MSE). For the train-test split, I divide the whole dataset on 76% or 363,719 examples for training and the remaining 34% or 162,717 examples for testing. Furthermore, I use the Stratify method to split the train and test set to retain the class ratios between the dataset. I also provide the output after applying the variance threshold and ANOVA SelectK-Best feature selection methods in table 12. Additionally, there is also the Mean Squared Error (MSE) value in this table.

TABLE XII. THE OUTCOME OF TRAIN-TEST SPLIT, FEATURE SELECTION METHODS, AND MEAN SQUARED ERROR(MSE)

| Algorithm | Train-Test Split Accuracy % (76% train, 34% test) | ANOVA Select-K-Best Accuracy (%) | Variance Threshold Accuracy (%) | Mean Squared Error (MSE) |
|---|---|---|---|---|
| Random Forest | 90.9% | 82.7% | 44.34% | 7.65 |
| Decision Tree | 99.07% | 99.1% | 44.4% | 0.0090 |
| KNN | 76.8% | 77.0% | 36.5% | 8.85 |
| LDA | 73.3% | 73.3% | 44.2% | 7.26 |
| ET | 82.7% | 82.7% | 44.3% | 6.21 |
| ADA | 45.1% | 45.1% | 42.3% | 22.2 |
| Ensemble 1 | 90.0% | 90.0% | 44.3% | 3.69 |
| Ensemble 2 | 65.8% | 72.0% | 59.6% | 7.71 |
| Ensemble 3 | 82.6% | 82.7% | 44.3% | 6.89 |
| Ensemble 4 | 97.7% | 91.8% | 44.2% | 2.35 |

*c) Confusion Matrix, Precision, Recall, and F1-Score:*

In this part of this paper, among all the 10 aforementioned architectures, I show the value of confusion matrix, precision, recall, and F1-score of four selected architectures (because of the page limitations). To display the result here, I only choose one particular evaluation criterion from each selected model by which I get superior accuracy. For example, for Random Forest, I provide the confusion matrix, precision, recall, and F1-score of Random oversampling.

From figure 12 and table 13, we can see the details performance analysis of the random forest algorithm. For most of the classes, this model fits perfectly with this dataset. For class 0 and 7, the recall and f1 scores are significantly low.

TABLE XIII. PRECISION, RECALL, AND F1-SCORE OF RANDOM FOREST

| Class | Precision | Recall | F1-Score |
|---|---|---|---|
| 0.aggravated-assault (AA) | .99 | .28 | .44 |
| 1. all-other-crimes (AOC) | .95 | .98 | .97 |
| 2.arson (AR) | .92 | .92 | .92 |
| 3.auto-theft (AT) | .96 | .97 | .97 |
| 4.burglary (BL) | .73 | 1.00 | .84 |
| 5.drug-alcohol (DA) | .98 | .93 | .95 |
| 6. larceny (LA) | .76 | .99 | .86 |
| 7.murder (MU) | 1.00 | .01 | .02 |
| 8.other-crimes-against-persons (OCAP) | .68 | .96 | .80 |
| 9. public-disorder (PD) | .95 | .89 | .92 |
| 10.robbery (RO) | 1.00 | .38 | .55 |
| 11.sexual-assault (SA) | 1.00 | 1.00 | 1.00 |
| 12.theft-from-motor-vehicle (TFMV) | 1.00 | .64 | .78 |
| 13. traffic-accident (TA) | 1.00 | 1.00 | 1.00 |
| 14.white-collar-crime (WCC) | 1.00 | .11 | .19 |
| **Accuracy** | | | **.91** |
| **Macro Average** | **.87** | **.68** | **.69** |
| **Weighted Average** | **.93** | **.91** | **.90** |

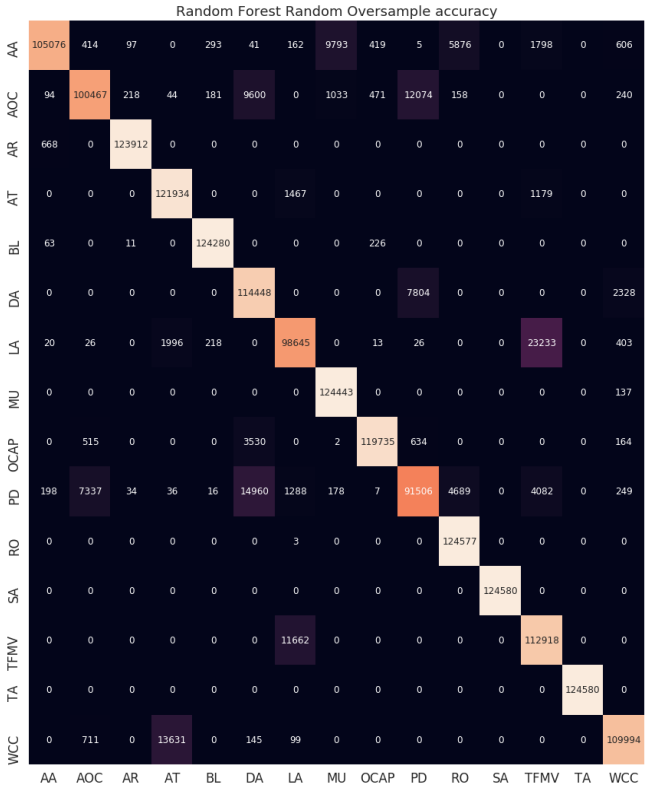

Fig. 12. Confusion Matrix of Random Forest

TABLE XIV. PRECISION, RECALL, AND F1-SCORE OF DECISION TREE

| Class | Precision | Recall | F1-Score |
|---|---|---|---|
| 0.aggravated-assault (AA) | 1.00 | 1.00 | 1.00 |
| 1. all-other-crimes (AOC) | 1.00 | 1.00 | 1.00 |
| 2.arson (AR) | 1.00 | 1.00 | 1.00 |
| 3.auto-theft (AT) | 1.00 | 1.00 | 1.00 |
| 4.burglary (BL) | 1.00 | 1.00 | 1.00 |
| 5.drug-alcohol (DA) | 1.00 | 1.00 | 1.00 |
| 6. larceny (LA) | 1.00 | 1.00 | 1.00 |
| 7.murder (MU) | 1.00 | 1.00 | 1.00 |
| 8.other-crimes-against-persons (OCAP) | 1.00 | 1.00 | 1.00 |
| 9. public-disorder (PD) | 1.00 | 1.00 | 1.00 |
| 10.robbery (RO) | 1.00 | 1.00 | 1.00 |
| 11.sexual-assault (SA) | 1.00 | 1.00 | 1.00 |
| 12.theft-from-motor-vehicle (TFMV) | 1.00 | 1.00 | 1.00 |
| 13. traffic-accident (TA) | 1.00 | 1.00 | 1.00 |
| 14.white-collar-crime (WCC) | 1.00 | 1.00 | 1.00 |
| **Accuracy** | **1.00** | **1.00** | **1.00** |
| **Macro Average** | **1.00** | **1.00** | **1.00** |
| **Weighted Average** | **1.00** | **1.00** | **1.00** |

From figure 13 and table 14, we can clearly see that the decision tree shows the overfitting tendency for this dataset.

TABLE XV. PRECISION, RECALL, AND F1-SCORE OF ENSEMBLE MODEL 1

| Class | Precision | Recall | F1-Score |
|---|---|---|---|
| 0.aggravated-assault (AA) | 0.88 | 0.64 | 0.74 |
| 1. all-other-crimes (AOC) | 0.95 | 0.99 | 0.97 |
| 2.arson (AR) | 1.00 | 0.23 | 0.37 |
| 3.auto-theft (AT) | 0.97 | 0.96 | 0.97 |
| 4.burglary (BL) | 0.82 | 1.00 | 0.90 |
| 5.drug-alcohol (DA) | 0.97 | 0.94 | 0.96 |
| 6. larceny (LA) | 0.63 | 1.00 | 0.77 |
| 7.murder (MU) | 1.00 | 0.09 | 0.16 |
| 8.other-crimes-against-persons (OCAP) | 0.84 | 0.96 | 0.89 |
| 9. public-disorder (PD) | 0.95 | 0.80 | 0.87 |
| 10.robbery (RO) | 1.00 | 1.00 | 1.00 |
| 11.sexual-assault (SA) | 1.00 | 1.00 | 1.00 |
| 12.theft-from-motor-vehicle (TFMV) | 1.00 | 0.36 | 0.53 |
| 13. traffic-accident (TA) | 1.00 | 1.00 | 1.00 |
| 14.white-collar-crime (WCC) | 1.00 | 0.14 | 0.25 |
| **Accuracy** | | | **0.92** |
| **Macro Average** | **0.93** | **0.74** | **0.76** |
| **Weighted Average** | **0.92** | **0.90** | **0.89** |

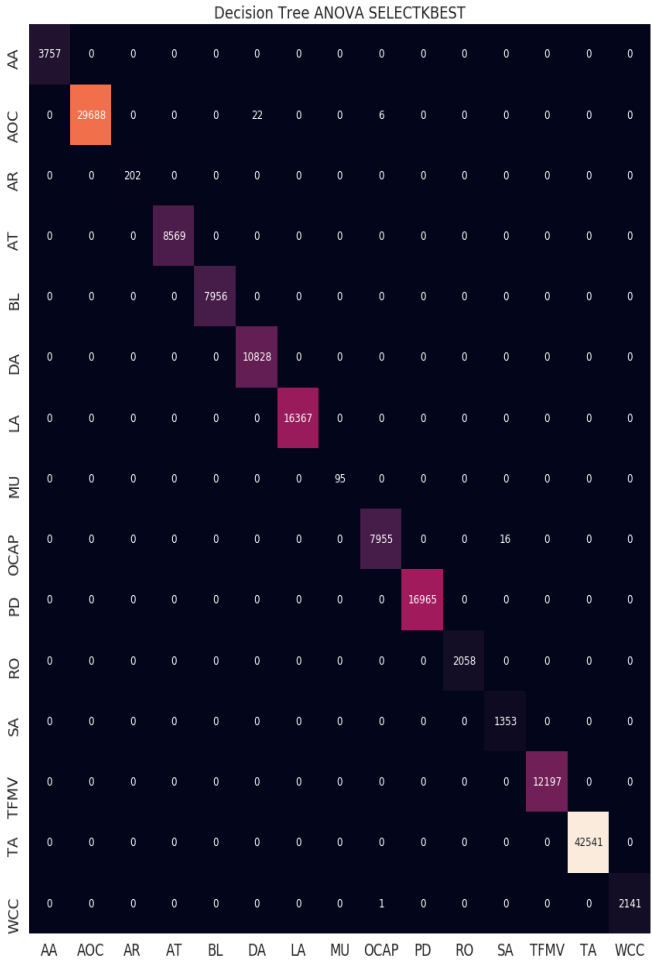

Fig. 13. Confusion Matrix of Decision Tree

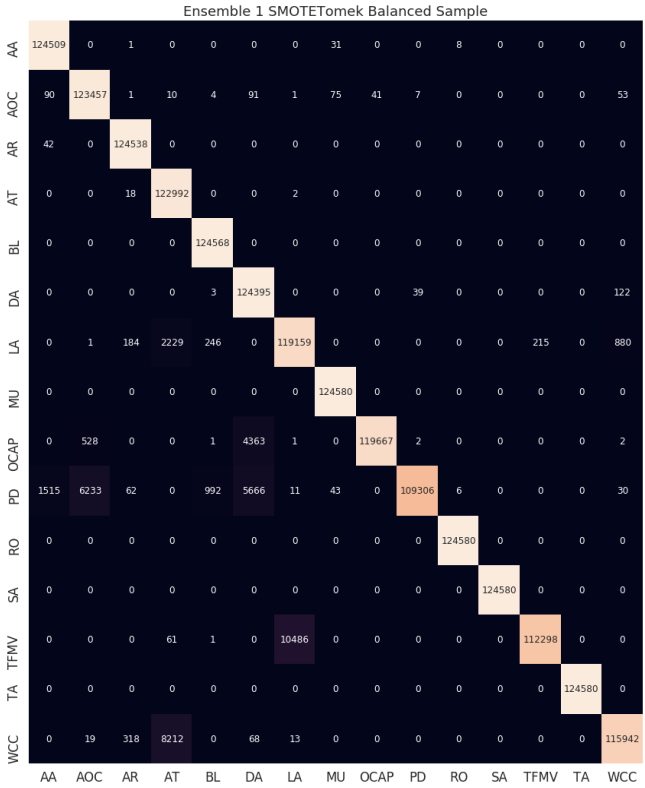

Fig. 14. Confusion Matrix of Ensemble Model 1

In figure 14 and table 15, I show the performance of the ensemble model 1. The recall value of the category murder and arson is notably low than the other attributes. Overall this model actually provides me a good prediction accuracy.

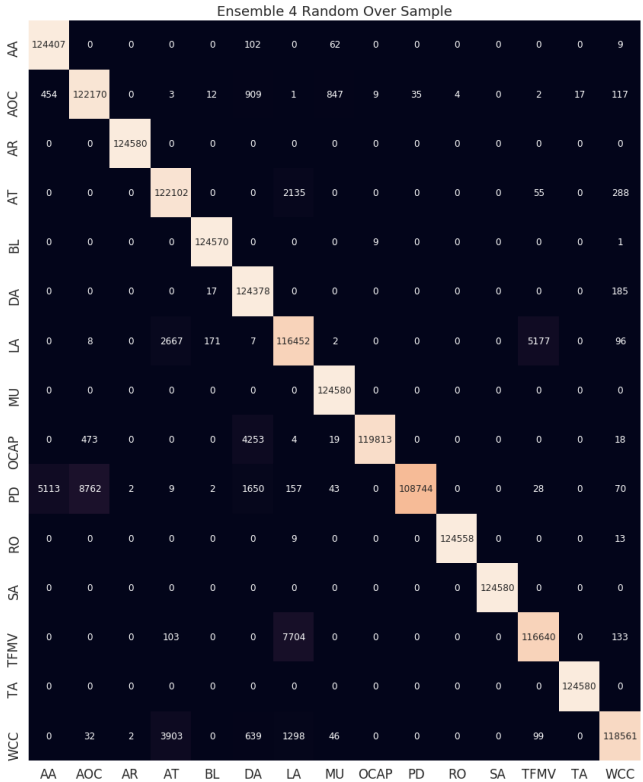

Fig. 15. Confusion Matrix of Ensemble Model 4

TABLE XVI. PRECISION, RECALL, AND F1-SCORE OF ENSEMBLE MODEL 4

| Class | Precision | Recall | F1-Score |
|---|---|---|---|
| 0.aggravated-assault (AA) | .98 | .97 | .98 |
| 1. all-other-crimes (AOC) | .95 | 1.00 | .97 |
| 2.arson (AR) | 1.00 | .55 | .71 |
| 3.auto-theft (AT) | .99 | .98 | .99 |
| 4.burglary (BL) | 1.00 | 1.00 | 1.00 |
| 5.drug-alcohol (DA) | .98 | 1.00 | .99 |
| 6. larceny (LA) | .91 | 1.00 | .95 |
| 7.murder (MU) | .88 | .88 | .88 |
| 8.other-crimes-against-persons (OCAP) | .98 | .96 | .97 |
| 9. public-disorder (PD) | 1.00 | .93 | .97 |
| 10.robbery (RO) | 1.00 | 1.00 | 1.00 |
| 11.sexual-assault (SA) | 1.00 | 1.00 | 1.00 |
| 12.theft-from-motor-vehicle (TFMV) | 1.00 | .90 | .95 |
| 13. traffic-accident (TA) | 1.00 | 1.00 | 1.00 |
| 14.white-collar-crime (WCC) | 1.00 | .79 | .88 |
| **Accuracy** | | | **.98** |
| **Macro Average** | **.92** | **.87** | **.89** |
| **Weighted Average** | **.98** | **.98** | **.98** |

From figure 15 and Table 16, I can reach a decision that based on accuracy, and the performance Ensemble Model 4 gives me a perfect outcome for this dataset. Furhtmeore, it generalizes the dataset ideally for every category of crimes.

*d) Paired-T Test:* By the paired-t-test, I want to inspect that the four selected algorithms are significantly different than each other or not. Moreover, I also want to see that these models can able to reject the null hypothesis or not. For the first T-test, I choose a random forest and decision tree. For the second one, I select Ensemble Model 1 and Model 4.

TABLE XVII. PAIRED-T-TEST FOR THE FOUR SELECTED MODELS

| Algorithms | P-Value | T-Statistics | Decision |
|---|---|---|---|
| N1= Random Forest N2= Decision Tree | 0.002 | 4.47 | P<T: cannot reject the null hypothesis and may conclude that the performance of the two algorithms is significantly different |
| N1= Ensemble Model 1 N2= Ensemble Model 4 | 4.26 | 1.39 | P>T: can reject the null hypothesis and may conclude that the performance of the two algorithms is not significantly different |

*e) ROC Curve:* Here, I provide the ROC curve for Random Forest, Decision Tree, LDA, and KNN. Through, ROC Curve, I want to show the ranking of each class of crimes.

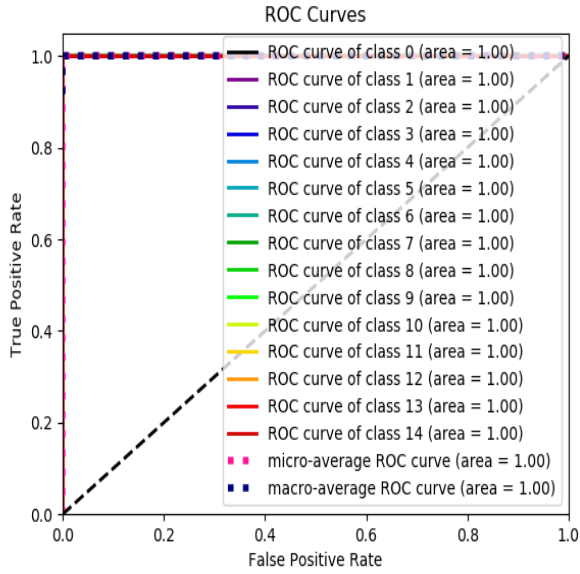

Fig. 16. ROC Curve for Decision Tree

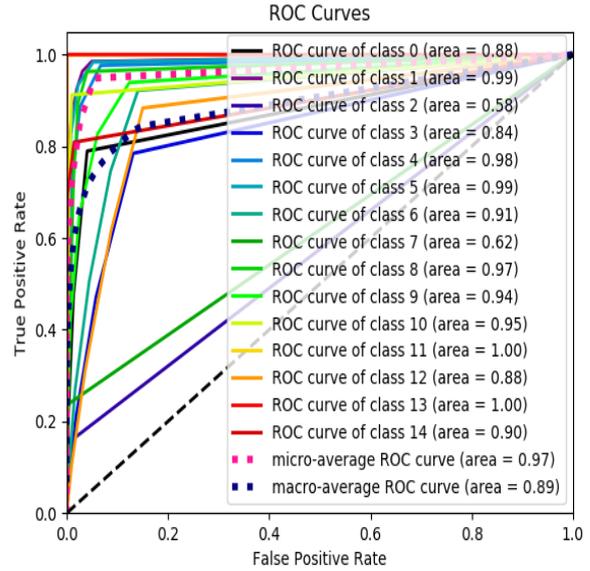

Fig 19. ROC Curve for KNN

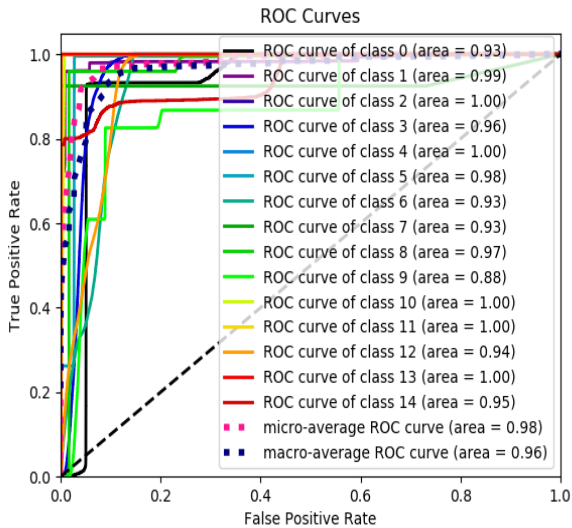

Fig. 17. ROC Curve for LDA

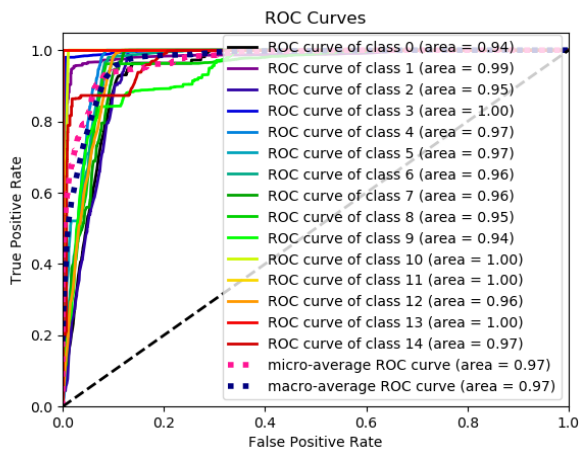

Fig. 18. ROC Curve for Random Forest

VI. CONCLUSION AND FUTURE WORK

Crime is a massive problem all over the world, which can damage our economy as well as social growth. The main goal behind this project is to provide an automated system that can predict the crime types based on diverse attributes. I believe this kind of crime prediction system always helps police, law enforcement authorities, security agencies, and the public to forecast the real crime incidents in advance. Besides, crime prediction and data mining tasks from the real-world dataset is an important task, but at the same time, it is a challenging one too. Most of the dataset have miscellaneous information and also containing many noisy data. To pick the key attributes from these various information is one of the most difficult issues. Therefore, efficient data collection, data mining, and accurate classification algorithm apply can be a crucial part of these kinds of projects.

Here, I apply many data visualization and data mining strategies on the Denver crime dataset to reveling many useful information. Before beginning the classification activities, I generated many graphs to inspect and experiment with several statistical details, which give me better intuition about this dataset. Then I applied many data pre-processing techniques and several classification algorithms to analyze the outcomes based many popular evaluation techniques. Among these algorithms, six are popular classification models, and the other four are ensembles models. Except for the AdaBoost classifier, most of the algorithms yield adequate accuracy, and some of them give me more than 90% accuracy in every evaluation criterion.

Furthermore, to accomplish the details performance analysis, I check the confusion matrix, precision, recall, F1-score, paired-T-test, and ROC curve for several models. I find many shortcomings of some algorithms, and on the other hand, some are executed really superior results. Overall, Ensemble Model 4 classifies 15 different categories of crimes with more than 90% accuracy in every evaluation method and provides an excellent performance analysis outcome. Thus, I am hopeful that maybe after a few improvements, it is possible to assemble a futuristic prediction system based on my proposed steps.

In the future, I want to solve several issues of this project. First of all, the decision tree gives me a strange output in every evaluation step. I tried many different combinations of parameters for this, but it is ended up being overfitted to this dataset. I will employ several data transformations techniques in the future to solve the complexities of this algorithm. Secondly, many algorithms produce a considerable amount of time complexities to accomplish their predictions, which I think is not suitable for any practical life application. Moreover, I am not able to implement SVM because of this time-related issue. So, I will try to find some efficient way to solve this matter. Lastly, progress my work ahead I will build a Deep Neural Network to compare the performance with these traditional Machine Learning approaches. I will apply this dataset with many existing deep learning models in the upcoming time.